\documentclass{sig-alternate-05-2015}

\usepackage{color}
\usepackage{algorithm}
\usepackage{algorithmic}
\usepackage{bm}

\newcommand{\INPUT}{\item[\myinput]}
\newcommand{\myinput}{\textbf{Initialization:}}

\newcommand{\mywhile}{\textbf{repeat}}

\newcommand{\myendwhile}{\textbf{until}}

\begin{document}

\setcopyright{acmcopyright}

%
%
%

%

\title{Human Pose Estimation from Depth Images via \\
Inference Embedded Multi-task Learning}
%
%
%
%
%

\numberofauthors{1} 
%

\author{
\alignauthor
Keze Wang$^{1, 2}$, Shengfu Zhai$^{1, 2}$, Hui Cheng$^{1, 2}$\thanks{The corresponding author is Hui Cheng (Email: chengh9@mail.sysu.edu.cn). This work was supported in part by Guangdong Natural Science Foundation under Grant S2013050014548 and 2014A030313201, in part by State Key Development Program under Grant No 2016YFB1001000, and in part by the Fundamental Research Funds for the Central Universities. This work was also supported by Special Program for Applied Research on Super Computation of the NSFC-Guangdong Joint Fund (the second phase).}, Xiaodan Liang$^{1, 2}$, Liang Lin$^{1, 2}$ \\
\affaddr $^1$Sun Yat-Sen University, Guangzhou, China \\
$^2$Collaborative Innovation Center of High Performance Computing, \\
National University of Defense Technology, Changsha 410073, China
%
}

\CopyrightYear{2016}
\setcopyright{acmcopyright}
\conferenceinfo{MM '16,}{October 15-19, 2016, Amsterdam, Netherlands}
\isbn{978-1-4503-3603-1/16/10}\acmPrice{\$15.00}
\doi{http://dx.doi.org/10.1145/2964284.2964322}

\maketitle
\begin{abstract}
Human pose estimation (i.e., locating the body parts / joints of a person) is a fundamental problem in human-computer interaction and multimedia applications. Significant progress has been made based on the development of depth sensors, i.e., accessible human pose prediction from still depth images~\cite{rf12pami}. However, most of the existing approaches to this problem involve several components/models that are independently designed and optimized, leading to suboptimal performances. In this paper, we propose a novel inference-embedded multi-task learning framework for predicting human pose from still depth images, which is implemented with a deep architecture of neural networks. Specifically, we handle two cascaded tasks: i) generating the heat (confidence) maps of body parts via a fully convolutional network (FCN); ii) seeking the optimal configuration of body parts based on the detected body part proposals via an inference built-in MatchNet~\cite{mn15cvpr}, which measures the appearance and geometric kinematic compatibility of body parts and embodies the dynamic programming inference as an extra network layer. These two tasks are jointly optimized. Our extensive experiments show that the proposed deep model significantly improves the accuracy of human pose estimation over other several state-of-the-art methods or SDKs. We also release a large-scale dataset for comparison, which includes 100K depth images under challenging scenarios.
\end{abstract}

%
%


\section{Introduction}
\begin{figure}[!htb]
\centering
\includegraphics[width = 1\columnwidth]{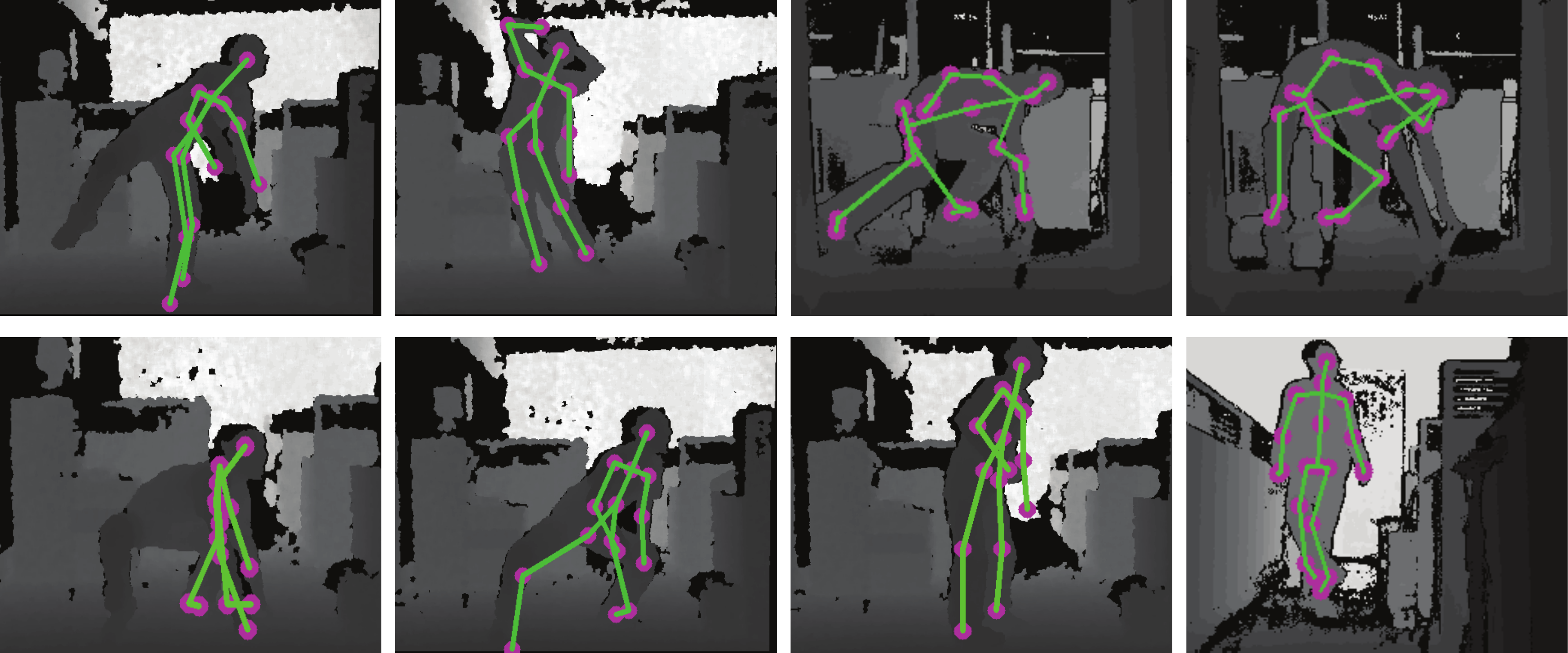}
\caption{Several failure examples generated by the latest version of Microsoft Kinect SDK in dealing with human pose estimation from depth data under challenging cases (e.g., unusual poses and self-occlusions).}\label{fig:motivation}
\end{figure}

Accurately locating the body parts/joints of a person (also referd as human pose estimation) is a widely studied problem in the research of multimedia. With the advent of cheap depth sensors like the structured-light scanner and ToF (Time of Flight) camera, the great performances have been achieved by the methods taking advantage of the depth data. For example, an approach based on decision forests has been proposed by researchers from Microsoft~\cite{rf11cvpr}, which classifies depth pixels corresponding to different parts of human body. This approach has been extended by Shotton et al.~\cite{rf12pami} by employing the regression forest. As a result, a great number of intelligent systems such as robotics and digital entertainments have been benefited a lot.

However, there are several difficulties under complex circumstances (e.g., the failure examples shown in Fig. 1):
\begin{enumerate}
 
\item \emph{Unusual poses/views.} The user may appear in diverse views or poses in real applications, while current solvers (e.g., the latest version of Microsoft Kinect SDK) often fail in handling exceptional and unusual cases. This limits the applications to subtle human-computer interaction, to some extent. 

\begin{figure}[!htb]
\centering
\includegraphics[width = 0.95 \columnwidth]{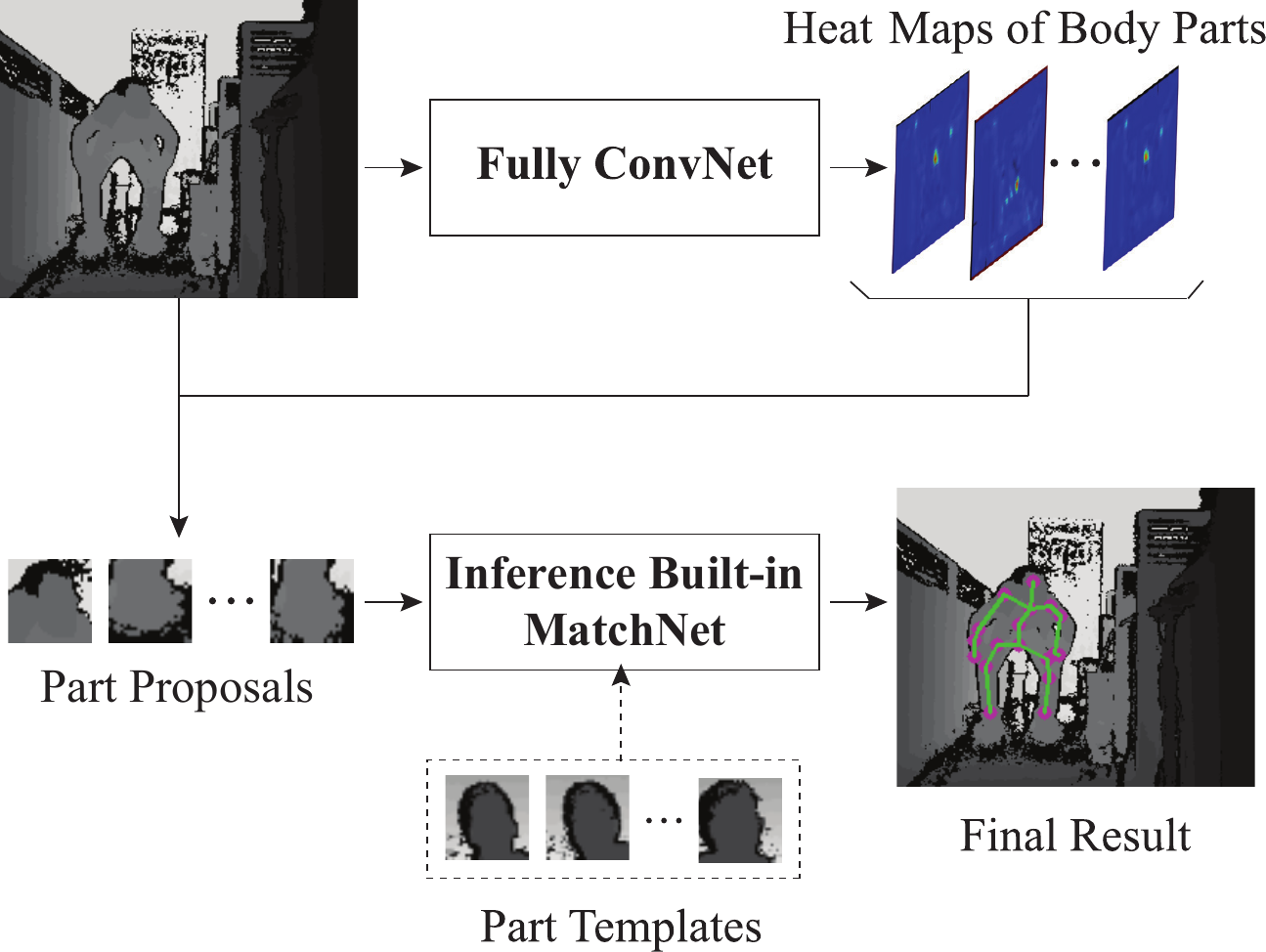}
\caption{Our framework subsequently performs two cascaded tasks. The first task predicts the heat maps of all body parts with a fully convolutional neural network. Thus the candidate proposals for all parts can be conveniently obtained. The second task infers the best configurations of body parts with a novel inference built-in MatchNet structure. This tailored MatchNet provides the appearance and geometric compatibility of body parts between the part proposals and part templates. The dynamic programming inference is naturally embedded into the network learning to generate final pose estimation results.}\label{fig:framework}
\end{figure} 

\item \emph{Self-occlusions.} The depth data captured from one single sensor inevitably contains much occlusions of the body parts, especially in playing complex gestures. The state-of-the-art performances are still far away from being satisfying. 
\end{enumerate}

The bottleneck of solving the above mentioned issues for further improving the accuracy performance is the fact that most of the traditional methods rely on handcrafted feature representations and carefully tuned constraints for modeling human body structures. And these methods usually involve several components/models that are independently designed and optimized. Recently proposed deep convolutional neural network (CNN) models have made incredible progress on image classification~\cite{alex12nips, keze2}, object recognition~\cite{alex12nips, rcnn14CVPR, dpl16cvpr, ruimao, xiaodan215iccv}, human activity recognition~\cite{scnn14acmmm, linlang2}, person re-indentification~\cite{guangrun, guangrun2, faqiang}, scene labeling~\cite{xiaodan2, xiaodan3, fullconv, sceneparsing1, sceneparsing2} and other vision problems. Improved performances are also achieved by the deep learning approaches~\cite{fp14eccv, jt14nips} on estimating human poses from RGB images. For example, Tompson et al.~\cite{jt14nips} proposed to impose a random field model upon CNNs for joint training and demonstrated the promising human pose estimation results on color data. However, directly applying these methods to the depth data is unfeasible due to the different challenges and requirements between the pose estimation tasks via color and depth data. Specifically, the depth images usually include sensor noises and preserve very coarse appearance details. Though this limits the human body part detection, the extra depth information can provide stronger contexts and skeletons for modeling human body structures and eliminating the effect of unrelated background clutters. Hence, we aim to develop a specialized neural network model that jointly incorporates the feature learning and the affinity constraints of body parts into a unified learning framework for the pose estimation from the depth data.

In this work, we propose a novel deep Inference-Embedded Multi-Task learning framework to estimate the human pose from still depth images. Specifically, our framework includes two networks for solving the two cascaded tasks, i.e., generating the heat (confidence) maps of body parts and further predicting body joints based on the generated body part proposals, respectively. Fig.~\ref{fig:framework} illustrates the overview architecture of our deep multi-task learning framework, in which these two tasks are conducted in the progressive coarse-to-fine manner and jointly optimized for pursuing higher pose estimation performance. 

The first task is performed by utilizing a fully convolutional network (FCN)~\cite{fullconv}, which generates heat maps for indicating the coarse locations of human body parts. The FCN enables regressing pixel-wise confidence maps of the body parts in depth images without the parameter-redundant fully-connected layers. With these maps, we can detect a batch of candidate patches (proposals) of human body parts for the subsequent task. 

Due to the low resolutions and discontinuities of depth images, the generated body part proposals may include many false positives. To overcome this issue in the second task, we develop an inference built-in MatchNet, which incorporates the feed-forward inference step into the deep architecture for seeking the optimal configuration of body joints. The original version of MatchNet~\cite{mn15cvpr} was proposed to jointly learn the patch-based features and distance metric. Here we adapt it for the following goals: i) measuring the unary appearance compatibility of the body part proposals with the standard part templates, and ii) measuring the geometric consistency of two body parts in the 3-D coordinate. We represent the human pose via a kinematic tree-based structure. (See Fig.~\ref{fig:motivation} for clarification)

Moreover, we treat the Dynamic Programming inference as an extra neural layer, which is built upon the neural layers to generate best part configurations based on calculated appearance and geometric compatibility. Given all the optimized parameters, the dynamic programming inference computes the optimal configuration from the leaf nodes (e.g., feet and hand nodes) of the proposed kinematic tree-based structure to the root node (say head nodes).

The two tasks in our learning framework are cascaded yet complementary to each other. The employed FCN is able to coarsely localize the candidate part proposals in a real-time speed. Then, the inference built-in MatchNet employs a more complex and deeper architecture to improve the final pose estimation accuracy from a global perspective. Therefore, jointly optimizing the two cascaded networks can benefit both accuracy and efficiency. Another advantage of our framework is the natural embedding of inference, that is, the heavy computational demand can be counter-balanced in a parallel manner by using Graphic Processing Unit (GPU).

\begin{figure*}[!htb]
\centering
\includegraphics[width =1 \textwidth]{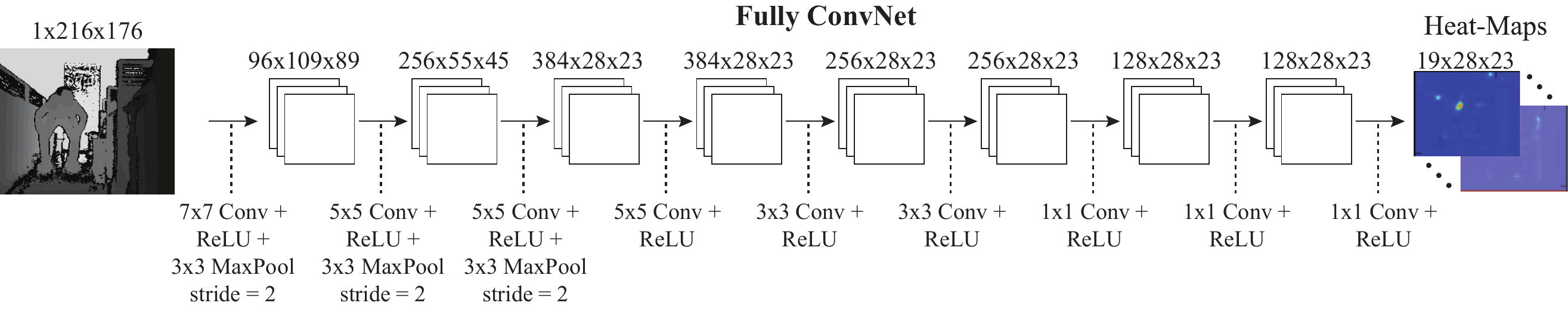}
\caption{The architecture of the proposed fully convolutional network for the part proposal generation.}\label{fig:fullconv}
\end{figure*}

The main \textbf{contributions} of this work are summarized as follows. First, we propose a task-driven deep learning model achieving a new state-of-the-art performance in predicting human body joints from still depth images. Our extensive experiments and comparisons with the existing approaches demonstrate the effectiveness and efficiency of our model. Second, the integration of dynamic programming inference with deep neural networks is original in the literature to the best of our knowledge. At last, we release a large-scale dataset of depth images for human pose estimation, which includes 100K depth images with various challenges.

The remainder of the paper is organized as follows. Sect.~\ref{sec:related} presents a review of related works. Then we present our inference embedded multi-task learning model in Sect.~\ref{sec:alg}, followed by the multi-task learning procedure of our model in Sect.~\ref{sec:learn}. The experimental results, comparisons and component analysis are exhibited in Sect.~\ref{sec:exper}. Sect.~\ref{sec:con} gives a conclusion of this paper.

\section{Related Work}
\label{sec:related}
Estimating the human pose from unconstrained color data is an important but challenging problem. Many approaches have been recently developed~\cite{cp80pami, shape02eccv, ia11cvpr, si03iccv, mengwang1, mengwang2, 14icme, xiaodan15iccv, xiaodan16cvpr, robust}. Mori et al.~\cite{shape02eccv} introduced an edge-based histogram, named ``shape-context'', to represent exemplar 2D views of the human body in a variety of different configurations and viewpoints with respect to the camera. Grauman et al.~\cite{si03iccv} introduced silhouette shape features to infer 3D structure parameters using a probabilistic multi-view shape model. More recently, Pictorial Structures~\cite{ps73tc, em00cvpr} and Deformable Part Models~\cite{dpm10pami, stc13acmmm, aom15pami} were proposed and have achieved tractable and practical performance on handling large variances of the body parts. As a result, large amounts of related models have been subsequently developed. Yang et al.~\cite{mof13pami} proposed to capture orientation, co-occurrence and spatial relations with a mixture of templates for each body part. In order to introduce richer high-level spatial relationships, Tian et al.~\cite{hmm12eccv} presented a hierarchical spatial model that can capture an exponential number of poses with a compact mixture representation on each part. Kiefel et al~\cite{fp14eccv} presented a binary conditional random field model to detect human body parts of articulated people in a single color image. 

As for estimating human pose from depth data, several approaches~\cite{rf11cvpr, hf11iccv, cr12cvpr, rf12pami, shotton13, rw15cvpr, 3d16cvpr, chu2016structure} have also been proposed. Shotton et al.~\cite{rf11cvpr} employed a random forest classifier to perform per-pixel classification of parts, and clustered these pixels for each part to localize the joints. Girshick et al.~\cite{hf11iccv} developed an approach for general-activity human pose estimation from depth images by building upon Hough forests. Shotton et al.~\cite{rf12pami} further made an extension by adopting the regression forest to directly regress the positions of body joints with the use of a large, realistic and highly varied synthetic set of training images. Jung et al.~\cite{rw15cvpr} proposed a highly efficient approach, which applies a supervised gradient descent and MCMC like random sampler in the form of Markov random walks beyond the pixel-wise classification. All these approaches suffer from the fact that they use hand crafted features such as HoG features, edges, contours, and color histograms.

Recently, deep convolutional neural networks (CNNs)~\cite{alex12nips} with sufficient training data have achieved remarkable success in computer vision. Several approaches have been proposed to employ CNNs to learn feature representation for human pose estimation. Toshev et al.~\cite{dp14cvpr} formulated the pose estimation as a deep CNN based regression problem towards body joints. Tompson et al.~\cite{jt14nips} proposed a novel hybrid architecture that consists of a convolutional neural network and a Markov Random Field for human pose estimation in the color monocular images. In~\cite{cn15cvpr}, Tompson further proposed an efficient `position refinement' model that is trained to estimate the joint offset location within a small region of the image. Chen et al.~\cite{idpr14nips} presented a deep graph model, which exploits deep CNNs to learn conditional probability for the presence of parts and their spatial relationships within these parts. 
\begin{figure*}[!htb]
\centering
\includegraphics[width = 0.85 \textwidth]{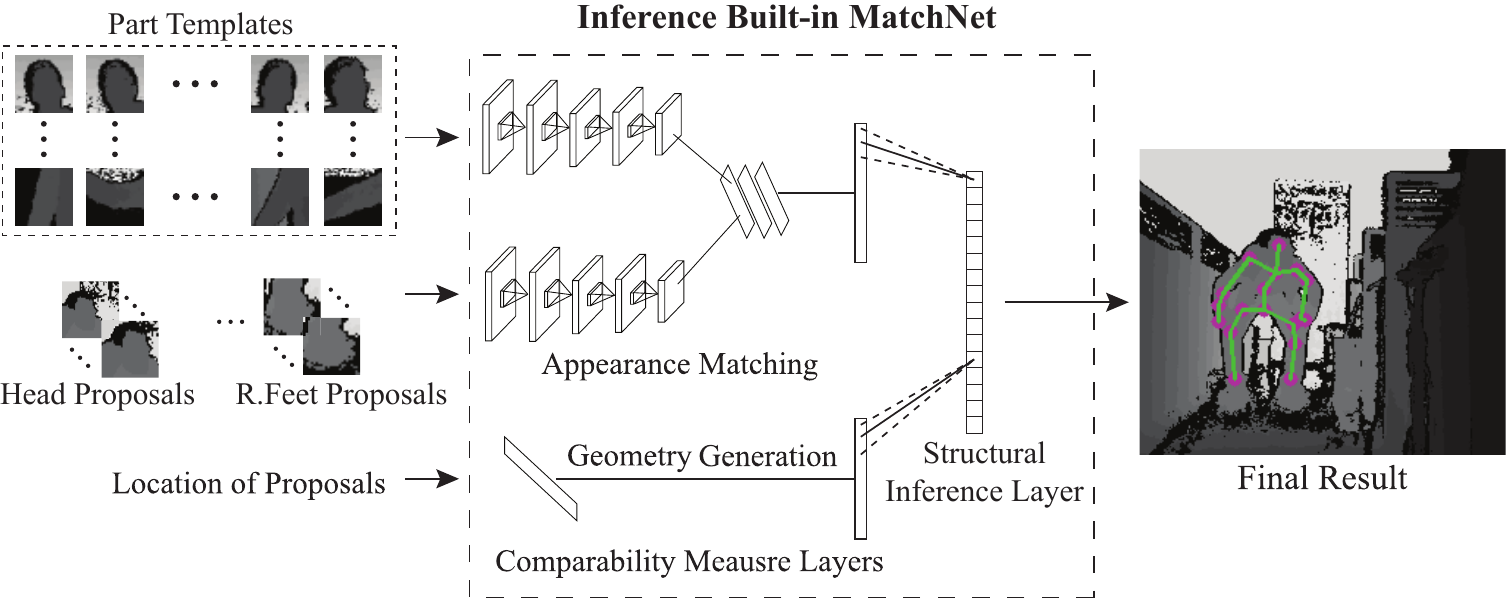}
\caption{The illustration of the inference built-in MatchNet for a fast part configuration inference.}\label{fig:matchnet}
\end{figure*}

\section{Inference Embedded Multi-task Learning}
\label{sec:alg}
In this section, we present our inference embedded multi-task learning framework. Our model first generates the heat (confidence) maps of body parts via a fully convolutional network, and further predicts body joints based on the generated body part proposals via the introduced inference built-in MatchNet.

\subsection{Fully Convolutional Network}
Motivated by the recent methods~\cite{jt14nips, dp14cvpr}, we regard the body part proposal generation as a problem of deep convolutional neural network regression. As shown in Fig.~\ref{fig:fullconv}, given an input depth image, our model produces a dense heat map output for each body part via a fully convolutional network (FCN), which is stacked by nine convolutional layers. Though a little deeper than~\cite{dp14cvpr} in the architecture, this FCN has smaller computation demands due to the small number and size of our used filters, i.e., much fewer network parameters. The output of FCN, called heat map, denotes a per-pixel likelihood for $K$ human body parts. We regard the coordinates of maximum value of the predicted heat-map $(x_k^*, y_k^*)$ as the center position of the $k$-th body part, $k=[1, ...,K]$.
\begin{equation}
(x_k^*, y_k^*) = \max_{x_k, y_k} z(k|I;\theta_c)
\end{equation}
where $\theta_c$ denotes the network parameters, $z(k|I;\theta_c)$ denotes the output heat map of the $k$-th body part for the input depth image $I$. Note that, since the neighboring pixels of the predicted heat-map also have similar heat values, only considering the maximum value may not achieve high accuracy for subtle localization of body parts. Based on this observation, we generate the predictions within the observation window (say $17\times17$) of the maximum value's position to increase the number of candidate body part proposals.

To express a body part proposal for the $k$-th body part, we specify the body part proposal $\textbf{b}_i^k$=[$x_i, y_i, w_i, h_i, k$], where ($x_i$, $y_i$) denotes the center coordinates of the body part proposal $\textbf{b}_i^k$, $w_i$ and $h_i$ denotes the width and height of $\textbf{b}_i^k$, respectively. Note that, since our final goal is the human pose estimation, $w_i$ and $h_i$ are not necessary to be very accurate, thus they are pre-defined and fixed as 10 pixels' distance between opposing joints on the human torso in our implementation. 

\subsection{Inference Built-in MatchNet} 
\label{sec:matchnet}
Our approach represents the human pose with a 3D kinematic $K$-node tree structure $\mathcal{G}$=\{$\mathcal{V}$, $\mathcal{E}$\}, where the node set $\mathcal{V}$ ($K$=$|\mathcal{V}|$) denotes the positions of body-joints/parts of the human pose, and the edge set $\mathcal{E}$ specifies the relationships of these body-joints/parts (see Fig.~\ref{fig:matchnet}). 
Given the candidates for the $k$-th and $m$-th body part in the depth image $I$, we define the 3D spatial relationship between the $i$-th and $j$-th body part proposals as $R(\textbf{b}_i^k, \textbf{b}_j^m|I)$. 
Considering the local appearance compatibility of each body part and global contextual relationships between body parts, the score function $F$ of our model is defined as the sum of unary and pairwise terms according to the proposed kinematic tree structure:
\begin{equation}
\label{equ:goal}
F(\hat{g}|I) = \sum_{k=1}^K U(\textbf{b}_i^k|I) + \sum_{m \in \Omega_k} R(\textbf{b}_i^k, \textbf{b}_j^m|I), 
\end{equation}
where $i, j \in \{1, ..., n\}$ and $n$ is the total number of the body part proposals, $I$ denotes the input depth image with $K$ body parts inside, the unary term $U(\textbf{b}_i^k|I)$ gives local evidence for the body part $k$ to lie at the center of the $i$-th body part proposal. The pairwise term $R(\textbf{b}_i^k, \textbf{b}_j^m|I)$ captures the geometric compatibility of the $k$-th and $m$-th body parts, where $m$ belongs to $k$'s neighboring body part set $\Omega_k$. $\hat{g}$ is the hidden variable in our model to configure the body part proposals for the final human pose estimation.

\textbf{Unary Terms:} The unary terms $\{U(\textbf{b}_i^k|I)\}_{k=1}^K$ are defined based on the local body part proposal $\textbf{b}_i^k$. Inspired by the recent deep patch-based matching technique~\cite{mn15cvpr}, we learn to extract features from body part proposals to calculate the appearance similarity. In our model, the appearance potential for the $i$-th body part proposal is measured according to the level of similarity between the extracted features of the proposal $\textbf{b}_i^k$ and the ones of the standard $k$-th body part templates, i.e., the similarity between input proposals $\textbf{b}_i^k$ and the part template set $\mathbf{B}_*^k$ = $\{\mathbf{b}_t^k\}_{t=1}^T$. To handle unusual poses/views, diverse templates for each body part are off-line learned via K-means clustering to represent various possible appearance/orientation of human poses. Specifically, the appearance potential score for the body part proposal $\mathbf{b}_i$ being the $i$-th body part is defined as:
\begin{equation}
\begin{split}
U(\mathbf{b}_i^k|I)=& \omega_k \cdot \phi(\mathbf{b}_i^k, \textbf{B}_*^k|I; \theta_u^k ) \\
=& \omega_k \cdot \sum_{t=1}^T \phi(\mathbf{b}_i^k, \mathbf{b}_t^k|I; \theta_u^k ),
\end{split}
\end{equation}
where $\phi(\cdot, \cdot|\cdot;\theta_u^k)$ is the appearance score term for being the $k$-th body part with the parameters $\theta_u^k$. 

\textbf{Pairwise Terms:} Besides matching the generated part proposal with part templates via the appearance similarity, we consider the geometry relationships among different human body parts, and define the pairwise terms accroding to the 3D kinematic constraint. Specifically, the constraints for the neighboring body parts are represented by the spatial deformation~\cite{dpm10pami} of the 3D geometric layout. The pairwise terms are:
\begin{equation}
R(\textbf{b}_i^k, \textbf{b}_j^m|I)= \Gamma_{km}\psi(\textbf{b}_i^k, \textbf{b}_j^m),
\end{equation}
where $\psi(\mathbf{b}_i^k, \mathbf{b}_j^m)$ denotes the 3D displacement vector, i.e., $[dx, dx^2, dy, dy^2, dz, dz^2]^T$ of the body part proposal $\mathbf{b}_j^m$ relative to $\mathbf{b}_i^k$, $\Gamma_{km}$ corresponds to the spatial deformation parameter, which indicates the contextual 3D spatial relations between the $k$-th and $m$-th body parts.

\begin{figure*}[!htb]
\centering
\includegraphics[width = 0.75 \textwidth]{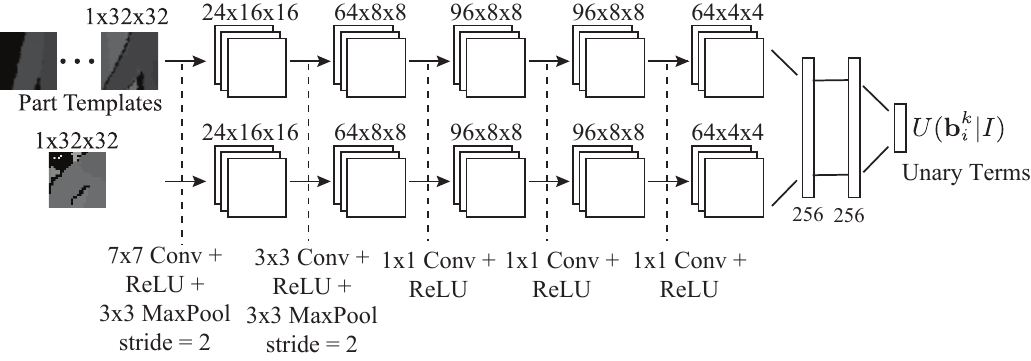}
\caption{The architecture of the appearance matching component.}\label{fig:amc}
\end{figure*} 

\textbf{Inference built-in MatchNet}
In our approach, the inference procedure is to predict body joints based on currently generated part proposals $\mathbf{B}=[..., \mathbf{b}_i^k, ...]$ by taking the appearance and geometric compatibility into account. The proposed inference built-in MatchNet integrates three components into a unified network architecture: i) Appearance matching to obtain appearance evidence (unary terms); ii) Geometry generating to calculate the 3D kinematic constraint (pairwise terms); iii) Structural inference to maximize the score function $F$ as: 
\begin{equation}
\max_{\hat{g}} \mathbf{w} \cdot F(\hat{g}|I),
\end{equation}
where $\mathbf{w}$ denotes all the parameters of the inference built-in MatchNet for simplicity, i.e., $\mathbf{w}$= $\{\Gamma_{km}, \omega_k\}_{k=1, m=1}^K$.

\section{Multi-task Learning}
\label{sec:learn}
Since our model includes two types of networks for human pose estimation, the network structure and parameters are optimized in a multi-task learning way. Let $\mathcal{D}=\{(I_1, g_1), ...,  (I_l, g_l), ...,(I_N, g_N)\}$, be a set of training samples, where $g_l$ denotes the annotated pose configuration for the $K$ body parts of the depth image $I_l$. In the following section, we will present the details of how our model learns to generate the part proposal $\textbf{b}_i^k$ and search for optimal configuration $\hat{g}_l$ from the candidate body part proposals under 3D kinematic constraint for the final pose estimation. 

\textbf{Fully Convolutional Network:}
We employ the widely used batch Stochastic Gradient Descent (SGD) algorithm to train this network. Specifically, given the input depth image $I_l$, we employ the Mean Squared Error (MSE) criterion to minimize the distance between the predicted output $z(k|I_l;\theta_c)$ and the ground truth heat-map $H(g_k|I_l)$ as~\cite{cn15cvpr}, where $H(g_k|I_l)$ is a 2D Gaussian with a small variance and mean centered at the ground truth joint locations:   
\begin{equation}
\label{equ:heatmap}
\min_{\theta_c} \sum_{l=1}^N \sum_{k=1}^K \Vert z(k|I_l; \theta_c) - H(g_k|I_l) \Vert^2_2.
\end{equation}
\textbf{Inference built-in MatchNet:}
During training the fully convolutional network, the inference built-in MatchNet is optimized simultaneously. As illustrated in Fig.~\ref{fig:matchnet}, there are two different types of layers in our model: the comparability measure layers and structural inference layer.

\emph{Comparability Measure Layers:}
The goal of these layers is to measure the appearance similarity between ground truth body parts and candidate body part proposals, which are generated by the fully convolutional network. Motivated by recent deep patch matching technique~\cite{mn15cvpr}, we employ a two-tower like neural network to jointly train the feature extraction and metric learning layer in a supervised setting (See Fig.~\ref{fig:amc}). The network parameters is also optimized by using SGD with the cross-entropy error:

\begin{equation}
\label{equ:am}
\begin{aligned}
E=-\frac{1}{N \cdot T} \sum_{l=1}^N \sum_{ t=1}^T \text{y}_i\log\left(\phi(\mathbf{b}_i^k, \mathbf{b}_t^k| I_l; \theta_u^k) \right) \\
+ ( 1 - \text{y}_i ) \log \left( 1 - \phi(\mathbf{b}_i^k, \mathbf{b}_t^k|I_l;\theta_u^k ) \right)
\end{aligned}
\end{equation}
where $\mathbf{b}_t^k$ is one of the body part templates for the $k$-th body part, and is obtained by clustering all the ground truth body part regions into $T$ centers via the K-means algorithm directly on the raw depth pixel. $\text{y}_i$ is the 0/1 label for the input proposal pair $\{ \mathbf{b}_i^k, \mathbf{b}_t^k\}$ (1 indicates match, i.e., the Intersection of Unit (IoU) between the predicted $\mathbf{b}_i^k$ and ground truth $\mathbf{b}_t^k$ is larger than 0.5). $\phi(\mathbf{b}_i^k, \mathbf{b}_t^k| I; \theta_u^k)$ is the output activation value of two-tower structure network, which represents the probability for being a matched pair. $\theta_u^k$ is the network parameter to be optimized for the $k$-th body part. 


\emph{Structural Inference Layer:} In our method, the configuration of body part proposals $\mathbf{B}$ for the depth image $I_l$ is determined by the latent variable $\hat{g}_l$. The MSE function $h(\hat{g}_l, g_l)$ is used to minimize the distance between the predicted body part location of the configuration $\hat{g}_l$ and the ground truth configuration of $g_l$. To simplify, $h(\hat{g}_l, g_l)$ is binarized according to the empirical threshold variable $\tau$ as follows:
\begin{equation}
h(\hat{g}_l, g_i) = \left\lbrace
\begin{aligned}
&0, ~~~\text{MSE}( \hat{g}_l, g_l ) \le \tau \\
&1, ~~~\text{otherwise.} \\
\end{aligned} \right.
\end{equation}
where MSE$( \hat{g}_l, g_l )$ denotes the average MSE error between the predict configurations $\hat{g}_l$ and the ground truth configuration $g_l$.

During the training phase, we optimize this layer by a latent learning method extended from the CCCP framework~\cite{cccp02nips} in a dynamic manner. The learning objective of this layer is to learn the parameter set $\mathbf{w}$ as described in Sect.~\ref{sec:matchnet}, which can be formulated as follows:

\begin{equation}
\label{equ:lsvm}
\begin{split}
\min_{\mathbf{w}} & \frac{1}{2} \Vert \mathbf{w} \Vert_2^2 + C \sum_{l=1}^N \max_{\hat{g}_l}( \mathbf{w} \cdot F( \hat{g}_l | I_l) - h( \hat{g}_l, g_l) \\
&~~~~~~ - \max_{\hat{g}_l} ( \mathbf{w} \cdot F( g_l | I_l) ) \\
& = \frac{1}{2} \Vert \mathbf{w} \Vert_2^2 + C \sum_{l=1}^N \max_{\hat{g}_l}( \mathbf{w} \cdot F( \hat{g}_l | I_l) - h( \hat{g}_l, g_l) \\
&~~~~~~ - C \sum_{l=1}^N \max_{\hat{g}_l} ( \mathbf{w} \cdot F( g_l | I_l) ) \\
&=f_1(\mathbf{w}) - f_2(\mathbf{w})
\end{split}
\end{equation}
where $\mathbf{w}=\{\omega_k$, $\Gamma_{km}\}_{k=1, m=1}^K$, and $C$ is a penalty parameter set empirically. Following the CCCP framework, we convert the function into a convex and concave form as $f_1(\mathbf{w}) - f_2(\mathbf{w})$, where $f_1(\mathbf{w})$ represents the first two terms, and $f_2(\mathbf{w})$ the last term. This leads to an iterative learning algorithm that alternates between estimating model parameters and the hidden variables. This type of learning problem is known as a structural SVM, and there are many solvers as the cutting plane solver~\cite{svmstruct} and SGD based solver~\cite{dpm10pami}. 

\begin{algorithm} [t]
\caption{}
\label{alg:alg_overview}
\begin{algorithmic}[1]
\REQUIRE Training samples $(I_1, g_1), (I_2, g_2), ..., (I_N, g_N)$
\ENSURE The parameters of our model \{$\theta_c, \theta_u \mathbf{w}$\}
\INPUT ~~\\
	 Initialize FCN's parameters $\theta_c$, the inference built-in MatchNet's network parameter $\theta_u$ and structural inference parameters $\mathbf{w}$.
	 
\mywhile	 

\STATE \ \ Optimize the fully convolutional network via Eqn.~(\ref{equ:heatmap})  \\	 
\ \ \ \ and generate candidate body part proposals $\textbf{B}$ ;
\STATE \ \ Given $\textbf{B}$, optimize the network parameter $\theta_u$ via \\
\ \ \ \ \ Eqn.~(\ref{equ:am}), and calculate the unary and pairwise terms;
\STATE \ \ Given the unary and pairwise terms, optimize \\ 
\ \ \ \ structural inference parameters $\mathbf{w}$ via Eqn.~(\ref{equ:lsvm});

\myendwhile \text{~The optimization function Eqn.~(\ref{equ:goal}) converges}
\end{algorithmic}
\end{algorithm}

\begin{figure}[t]
\centering
\includegraphics[width = 1 \columnwidth]{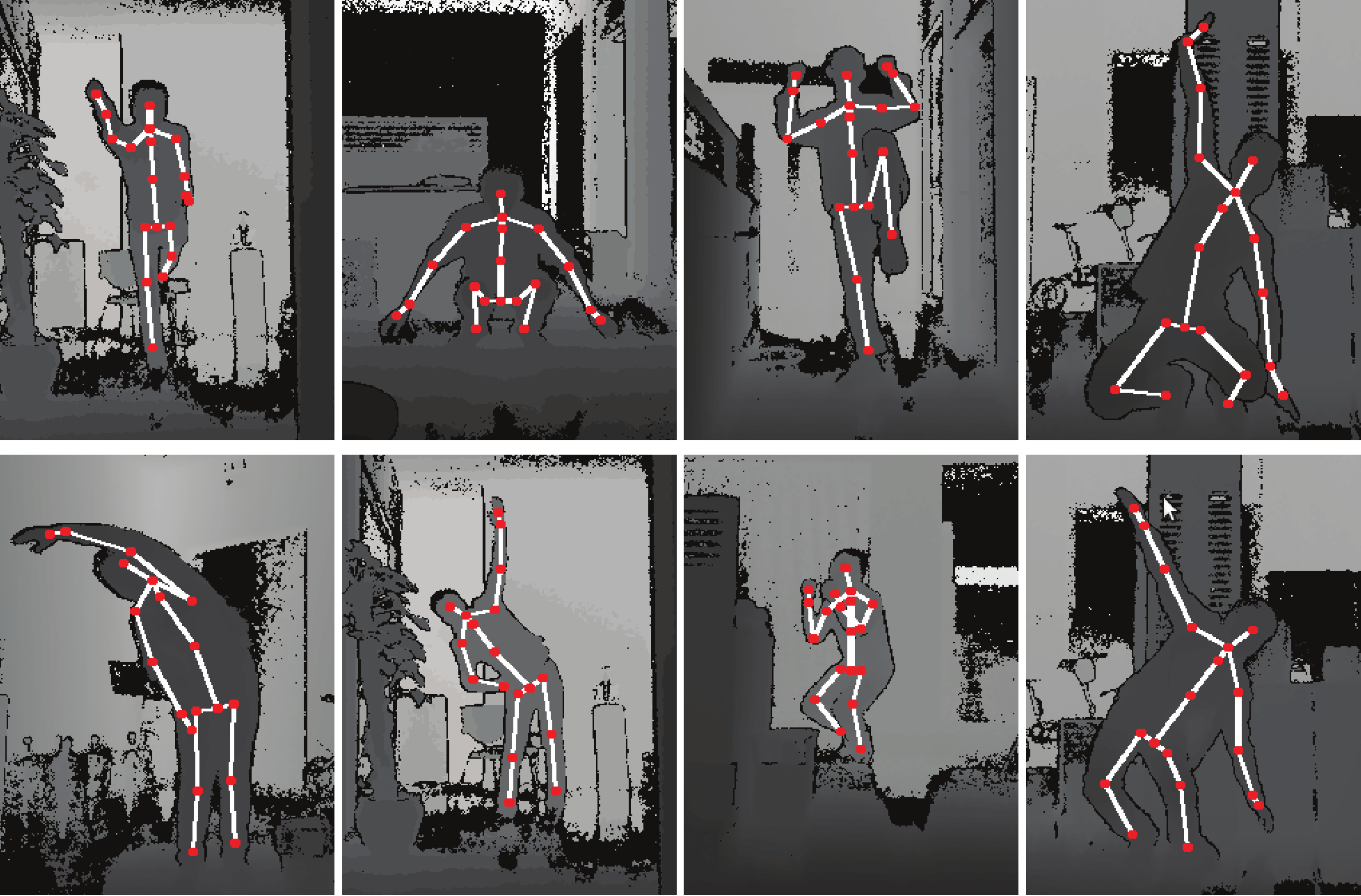}
\caption{Human pose examples from the K2HGD dataset. The ground truth body parts are illustrated in red and connected in white.}\label{fig:dataset}
\end{figure} 

The entire algorithm for multi-task learning can be summarized into Algorithm~\ref{alg:alg_overview}. 

\section{Experiments}
\label{sec:exper}
In this section, we compare our model with several state-of-the-art methods and SDKs, and justify the effectiveness of each component.

\subsection{Dataset and Parameter Setting}
\textbf{Dataset.}
We evaluate the human pose estimation performance of our inference-embedded multi-task learning framework on our Kinect2 Human Gesture Dataset (K2HGD)\footnote{http://vision.sysu.edu.cn/projects/k2hpd}, which includes about 100K depth images with various human poses under challenging scenarios. As illustrated in Fig.~\ref{fig:dataset}, this dataset consists of 19 body joints of 30 subjects under ten different challenging scenes. The subject is asked to perform both normal daily poses and unusal poses. The human body joints are defined as follows: \emph{Head, Neck, UpperSpine, MiddleSpine, DownSpine, RightShoulder, RightElbow, RightWrist, RightHand, LeftShoulder, LeftElbow, LeftWrist, LeftHand, RightHip, RightKnee, RightFoot, LeftHip, LeftKnee, LeftFoot}. The ground truth body joints are firstly estimated via the Kinect SDK, and further refined by active users. 

\begin{figure}[t]
\centering
\includegraphics[width = 0.85 \columnwidth]{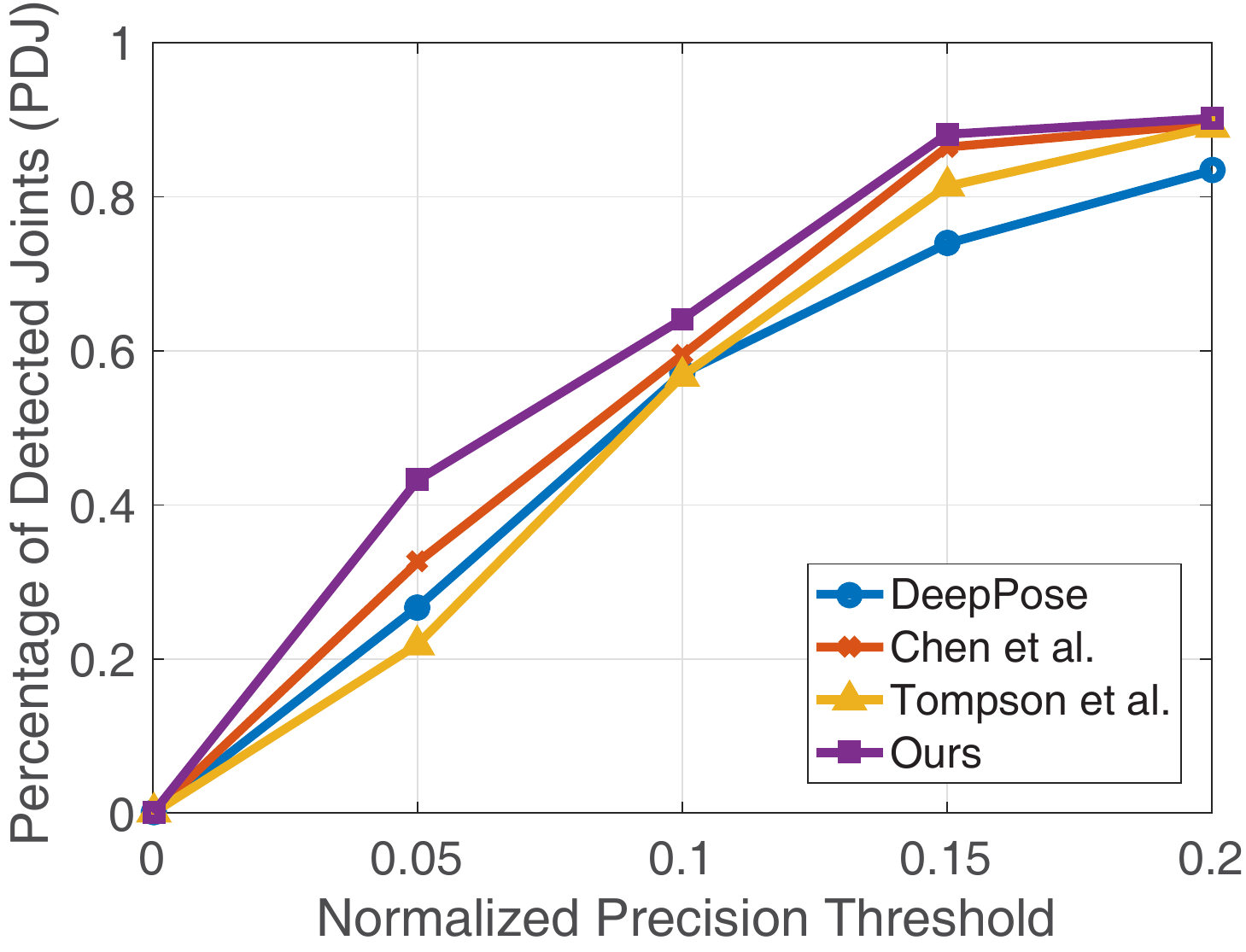}
\caption{Performance comparison with three state-of-the-art methods on PDJ curves of average body parts on the K2HGD dataset.}\label{fig:method}
\end{figure}

\begin{table}[t]
\scriptsize
\center
\setlength{\tabcolsep}{2.5pt}
\begin{tabular}{|c|c|c|c|c|}
\hline
\hline
Method & DeepPose & Chen et al. & Tompson et al. & Ours  \\
& \cite{dp14cvpr} &\cite{idpr14nips}&\cite{cn15cvpr} & \\
\hline
PDJ (0.05) & 26.8\% & 32.6\% & 21.9\% & \textbf{43.2}\% \\
PDJ (0.10) & 57.1\% & 59.5\% & 56.8\% & \textbf{64.1}\% \\
PDJ (0.15) & 73.9\% & 86.5\% & 81.3\% & \textbf{88.1}\% \\
PDJ (0.20) & 83.4\% & 89.5\% & 89.0\% & \textbf{91.0}\% \\
\hline
Average & 60.3\% & 67.0\% & 62.3\% & \textbf{71.6}\% \\
\hline
\hline
\end{tabular}
\vspace{1pt}
\caption{Detailed comparison of the estimation accuracy on the K2HGD dataset. The entries with the best precision values for each row are bold-faced.}\label{tab:detailed}
\end{table}

\begin{table}[t]
\scriptsize
\center
\setlength{\tabcolsep}{2.5pt}
\begin{tabular}{|c|c|c|c|c|}
\hline
\hline
Method & DeepPose & Chen et al. & Tompson et al. & Ours  \\
& \cite{dp14cvpr} &\cite{idpr14nips}&\cite{cn15cvpr} & \\
\hline
Time & 0.02 & 7.6 & 0.2 & 0.04 \\
\hline
\hline
\end{tabular}
\vspace{1pt}
\caption{Comparison of the average running time (seconds per image) on the K2HGD dataset.}\label{tab:time}
\end{table}

\textbf{Compared Methods.}
On the K2HGD dataset, we make a quantitative comparison with three deep state-of-the-art methods: DeepPose~\cite{dp14cvpr}, Tompson et al.~\cite{cn15cvpr} and Chen et al.~\cite{idpr14nips}. We adopt their public implementations. Due to commercial reasons, the widely used OpenNI SDK and Microsoft SDK are not supported to estimate human pose in given still depth images. Hence, we conduct a qualitative comparison with them. 

\begin{figure*}[!htb]
\centering
\includegraphics[width = 0.85 \textwidth]{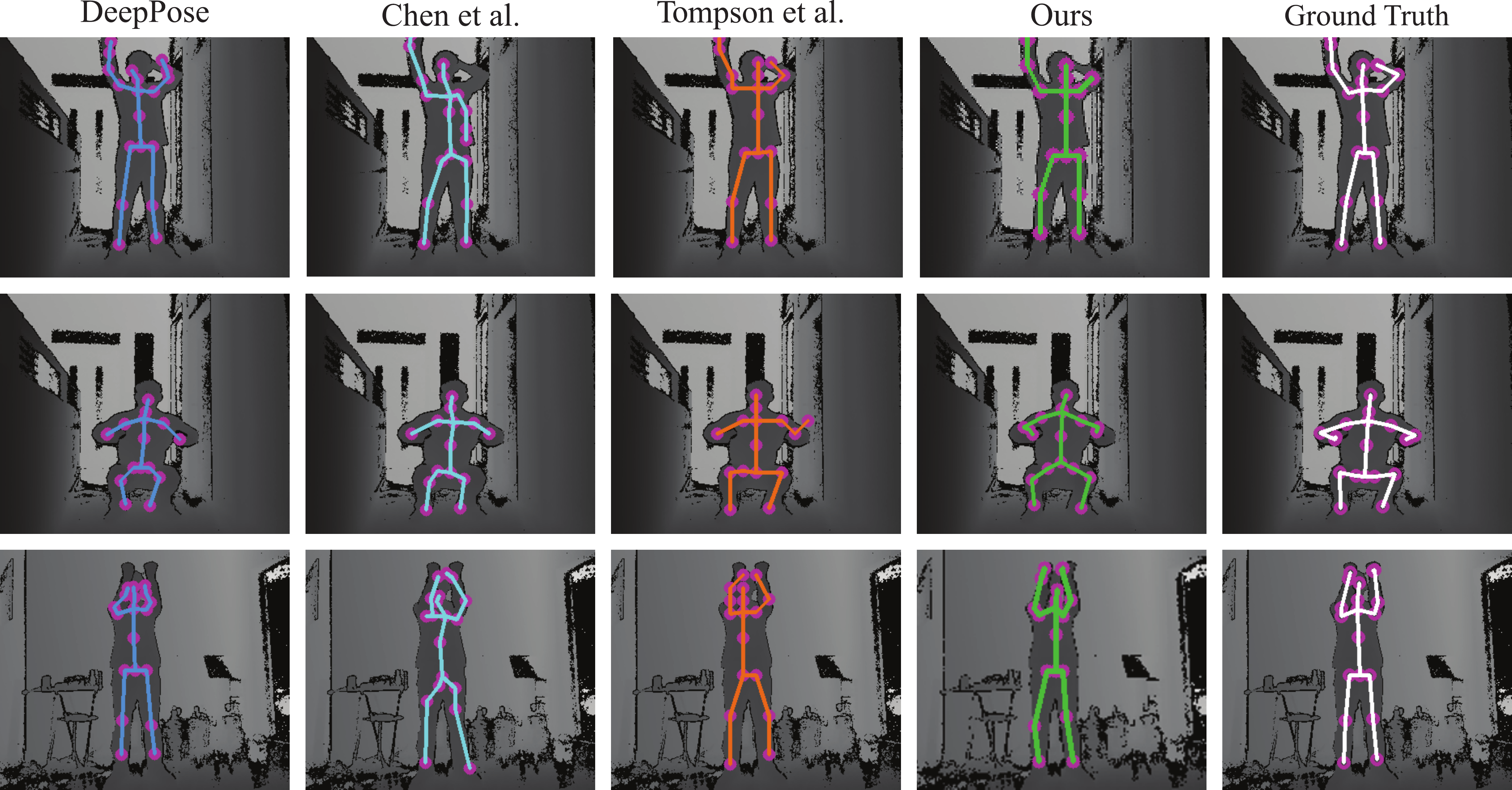}
\caption{Qualitative comparison with the existing state-of-the-art methods. The estimation is perfromed on the whole depth image. We have cropped some irrelevant background for better viewing.}\label{fig:visual2}
\end{figure*}

\begin{figure*}[!htb]
\centering
\includegraphics[width = 0.65 \textwidth]{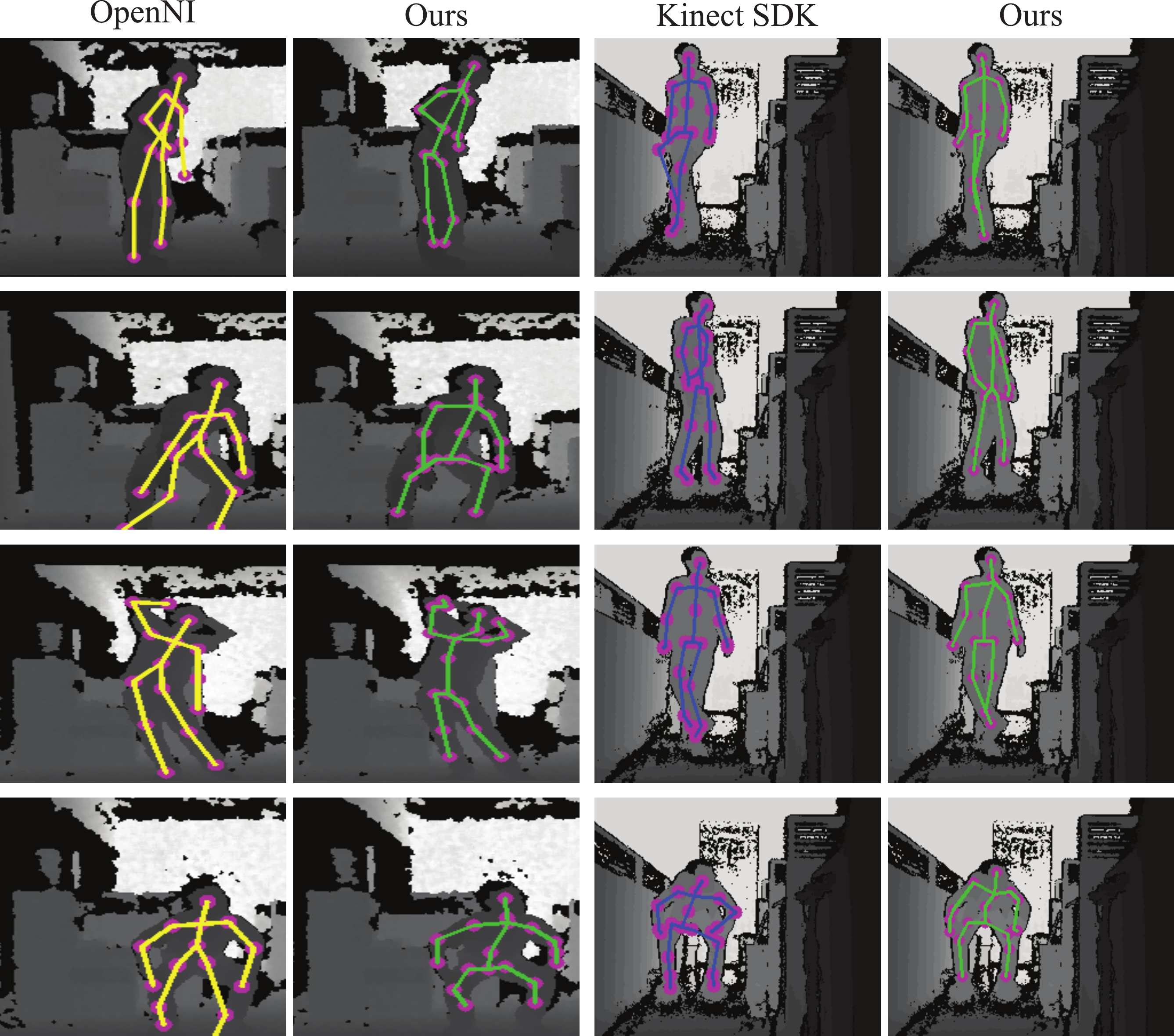}
\caption{Qualitative comparison with OpenNI and Kinect SDK. The results of OpenNI are in yellow lines. The results of Kinect SDK are in blue lines. The results of our method are in green lines.}\label{fig:visual}
\end{figure*}

\textbf{Metric.} To measure the accuracy of predicting human body joints, we employ the popular Percent of Detected Joints (PDJ) metric as the evaluation criterion. This metric considers a body joint is correctly estimated only if the distance between the predicted and the true joint position is within a certain fraction of the torso diameter. Given different normalized precision threshold (i.e., normalized distance to the ground truth body part), the estimation accuracy is obtained under different degrees of overall localization precision.

\textbf{Experimental Details.} 
In all experiments, 60K still depth images are used for training while the rest 40K used for testing. We set the parameters \{$n, \tau, K, C, T$\} = \{ 289, 0.2, 19, 0.001, 10\}. The patch size of each part is set to $32 \times 32$. Our model is implemented on Caffe library~\cite{caffe14}. All of our experiments are carried out on an intel 3.4 GHz and a NVIDIA GTX-980Ti GPU. In order to reduce overfitting, we employ the common cropping and horizontal-flipping strategy to augment the training data.  In this way, spatial relationships between body parts and their neighbors can be increased. The learning rate for the fully convolutional networks is 0.01 while the momentum is 0.9, the weight decay is 0.0005 and the batch size is 20. The number of body parts $K$ is set as 19 and 289 candidate proposals for each body part are generated. For the inference built-in MatchNet, the learning rate is 0.1 while the momentum is 0.9, the weight decay is 0.0005 and the batch size is 20.

\subsection{Results and Comparisons}
\textbf{Quantitative Comparison.} The quantitative comparison results is demonstrated in Fig.~\ref{fig:method} and Tab.~\ref{tab:detailed}. Fig.~\ref{fig:method} shows the PDJ curves of ours model over the other three compared state-of-the-art methods. As one can see that, our approach consistently outperforms all the compared methods under every normalized precision threshold, and are significantly superior to others under the requirement of strict estimation accuracy, i.e., the normalized precision threshold ranges from 0 to 0.1. As illustrated in Tab.~\ref{tab:detailed}, our model achieves 11.3\%, 4.6\%, 9.3\% better accuracy than DeepPose, the method of Chen et al. and Tompson et al. in average, respectively. 

To justify the effectiveness of the proposed inference built-in strategy, we further quantitatively analyze time efficiency of our method and all compared approaches on the K2HGD dataset. As demonstrated in the Tab.~\ref{tab:time}, our method runs 190x and 5x faster than the approach of Chen et al. and Tompson et al., respectively. Though our method contains the time-consuming inference step as Chen et al. and Tompson et al., inference in our method is embedded in the forward process of networks. Hence, our method is much faster. Since DeepPose directly regresses all body parts in a holistic manner from the depth image, our method is 2x slower than it. However, our performance is significantly better than DeepPose.

\textbf{Qualitative Comparison.} Fig.~\ref{fig:visual2} and Fig.~\ref{fig:visual} demonstrate the qualitative comparison of our method, OpenNI, latest Kinect SDK and three existing state-of-the-art methods. As one can see from Fig.~\ref{fig:visual} and Fig.~\ref{fig:visual2}, our method achieves more robust and accurate performance for predict unusual and self-occlusion poses.

\subsection{Empirical Analysis}
\begin{figure}[t]
\centering
\includegraphics[width = 0.85 \columnwidth]{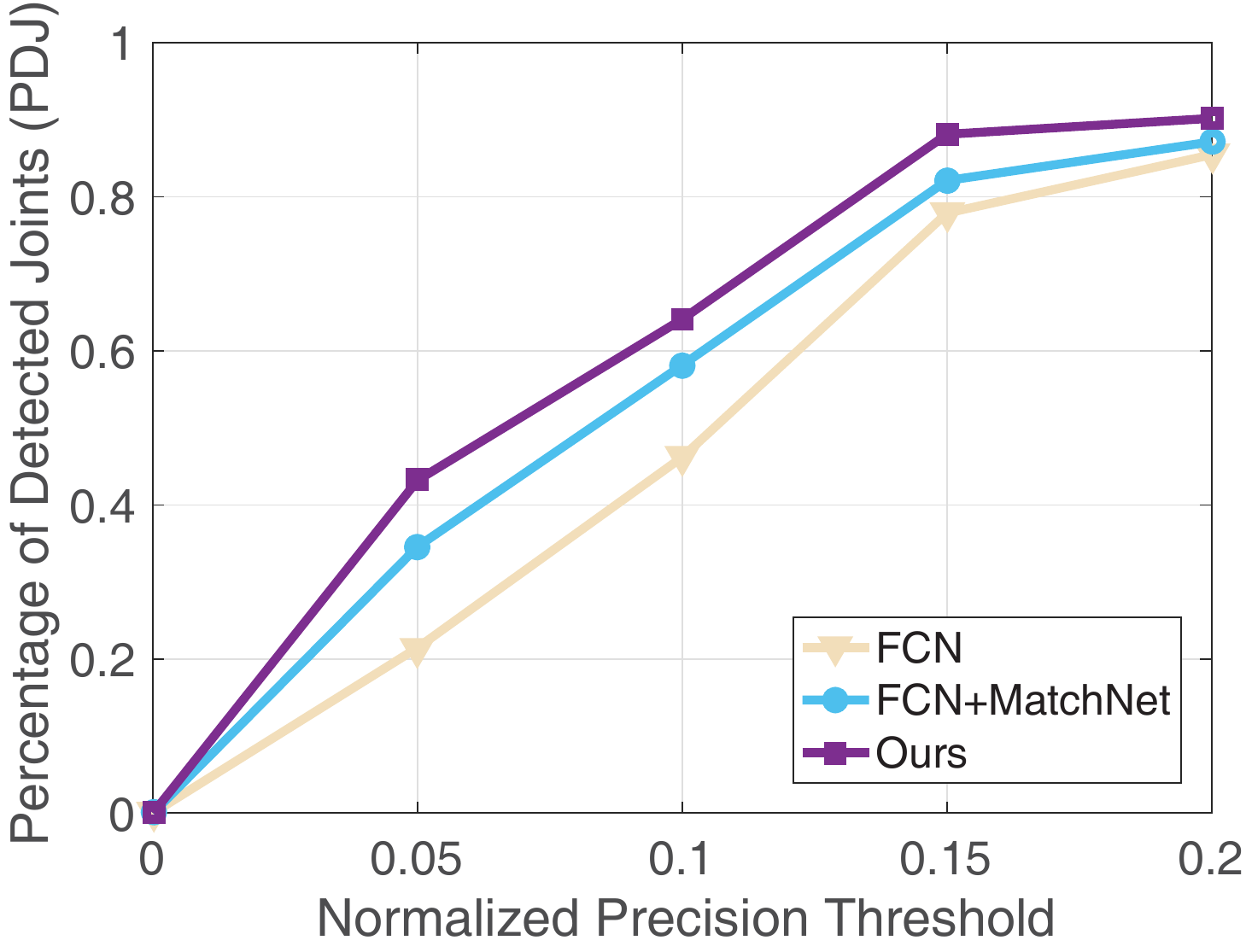}
\caption{Component analysis for the contribution of different network in our model on the K2HGD dataset. The PDJ curves imply the average pose estimation accuracy of all body parts.}\label{fig:component_curve}
\end{figure}

\begin{figure}[t]
\centering
\includegraphics[width = 0.95 \columnwidth]{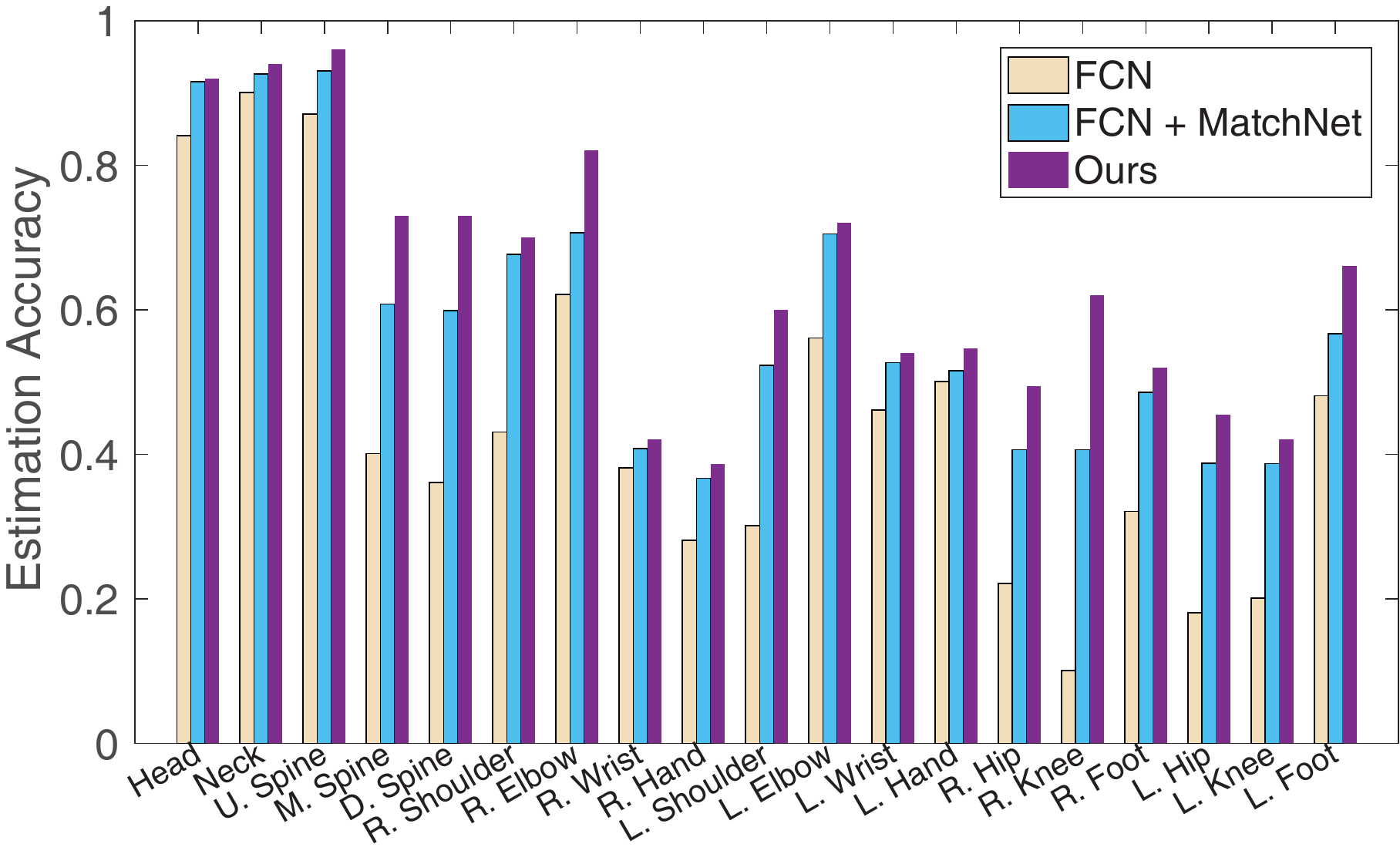}
\caption{Component analysis for the contribution of different network in our model on the K2HGD dataset. The average PDJ accuracy for each body part is reported.}\label{fig:component_bar}
\end{figure}

To perform a component analysis of our method, we conduct the following experiment to validate the contributions of different parts of the proposed framework. To justify the effectiveness of the proposed inference built-in MatchNet, we directly use the output, i.e., heat map, of the full convolutional network (FCN) inside our method to estimate human pose. This variant version of our method represents the pure performance of the FCN, and is denoted as ``FCN''; we also use the output of appearance matching component to predict human pose. This variant version of our method represents the combination of FCN and appearance evidence via the standard MatchNet, and is denoted as ``FCN+MatchNet''.
We compare it with our method both in average PDJ curves of body parts (See Fig.~\ref{fig:component_curve}) and average PDJ estimation accuracy for each body part. (See Fig.~\ref{fig:component_bar}). 

Fig.~\ref{fig:component_curve} demonstrates that there lies a significant performance gap between our method and FCN. This validates the superiority of our proposed inference built-in MatchNet. As one can see from Fig.~\ref{fig:component_curve}, ``FCN+MatchNet'' performs consistently better than FCN. This justifies the contribution of the appearance matching (unary term) inside our model. Moreover, we can also show the effectiveness of the 3D kinematic constraint (pairwise term) for the result that our method also outperforms FCN+MatchNet.

To explore the detailed improvement of our model's component, we report the PDJ for each body part in the Fig.~\ref{fig:component_bar}, which also shows that the proposed inference MatchNet effectively improve the estimation accuracy, especially for \{\emph{Elbow, Shoulder, Knee, Foot}\} body parts. 

%

\subsection{Limitations}
The proposed framework may fail to estimate the poses of multiple persons from the still depth image. As the employed fully convolutional network can only output 19 heat maps for the body parts of a single person, our model may generate unsatisfying estimations due to the confusion body parts caused by multiple persons. Another limitation stems from our assumption, i.e., most of main body parts (e.g., torso, shoulder and spin) should be visible from depth data. Based on this assumption, we propose the above mentioned 3D kinematic constraint for the final pose estimation. Hence, our model may not work well when all main body parts are severely occluded.

\section{Conclusion}
\label{sec:con}
We proposed a novel deep inference-embedded multi-task learning framework for predicting human pose from depth data. In this framework, two cascaded tasks were jointly optimized: i) detecting a batch of proposals of body parts via a fully convolutional network (FCN); ii) searching for the optimal configuration of body parts based on the body part proposals via a fast inference step (i.e., dynamic programming), which operates with a inference built-in MatchNet to incorporate the single term of appearance cost and the pairwise 3D kinematic constraint. We believe that it is general to be extended into other similar tasks such as object parsing. Extensive experiments validated the effectiveness of the proposed framework. In the future, we will try to support the pose estimation for multiple persons from depth data.

\section{Acknowledgments}
We would like to thank Huitian Zheng and Dengke Dong for their preliminary contributions on this project. We gratefully acknowledge the support of NVIDIA Corporation with the donation of the GPU used for this research.

%
\bibliographystyle{abbrv}
\bibliography{mybib}  
\end{document}